\documentclass[runningheads]{llncs}
\usepackage{graphicx}
\usepackage{color}
\usepackage{subcaption}
\newcommand{\BOS}{$\langle${\tt BOS}$\rangle$}
\newcommand{\EOS}{$\langle${\tt EOS}$\rangle$}
\newcommand{\PAD}{$\langle${\tt PAD}$\rangle$}
\newcommand{\CLS}{$\langle${\tt CLS}$\rangle$}
\begin{document}
%https://www.overleaf.com/project/61da073fee20a7ba88397b48
\title{Font Shape-to-Impression Translation}
\author{Masaya Ueda\inst{1} \and
Akisato Kimura\inst{2} \and
Seiichi Uchida\inst{1}}
\authorrunning{M. Ueda et al.}
\institute{Kyushu University, Fukuoka, Japan \and   
NTT Communication Science Laboratories, NTT Corporation, Japan}
\maketitle              % typeset the header of the contribution
\begin{abstract}
%The abstract should briefly summarize the contents of the paper in 15--250 words.
Different fonts have different impressions, such as elegant, scary, and cool. This paper tackles part-based shape-impression analysis based on the Transformer architecture, which is able to handle the correlation among local parts by its self-attention mechanism. This ability will reveal how combinations of local parts realize a specific impression of a font. The versatility of Transformer allows us to realize two very different approaches for the analysis, i.e., multi-label classification and translation. A quantitative evaluation shows that our Transformer-based approaches estimate the font impressions from a set of local parts more accurately than other approaches. A qualitative evaluation then indicates the important local parts for a specific impression. %
\end{abstract}

% =======================================================================
\section{Introduction\label{sec:intro}}

%---------------------------------
\begin{figure}[t]
    \centering
%    \vskip 5cm
    \includegraphics[width=1.0\linewidth]{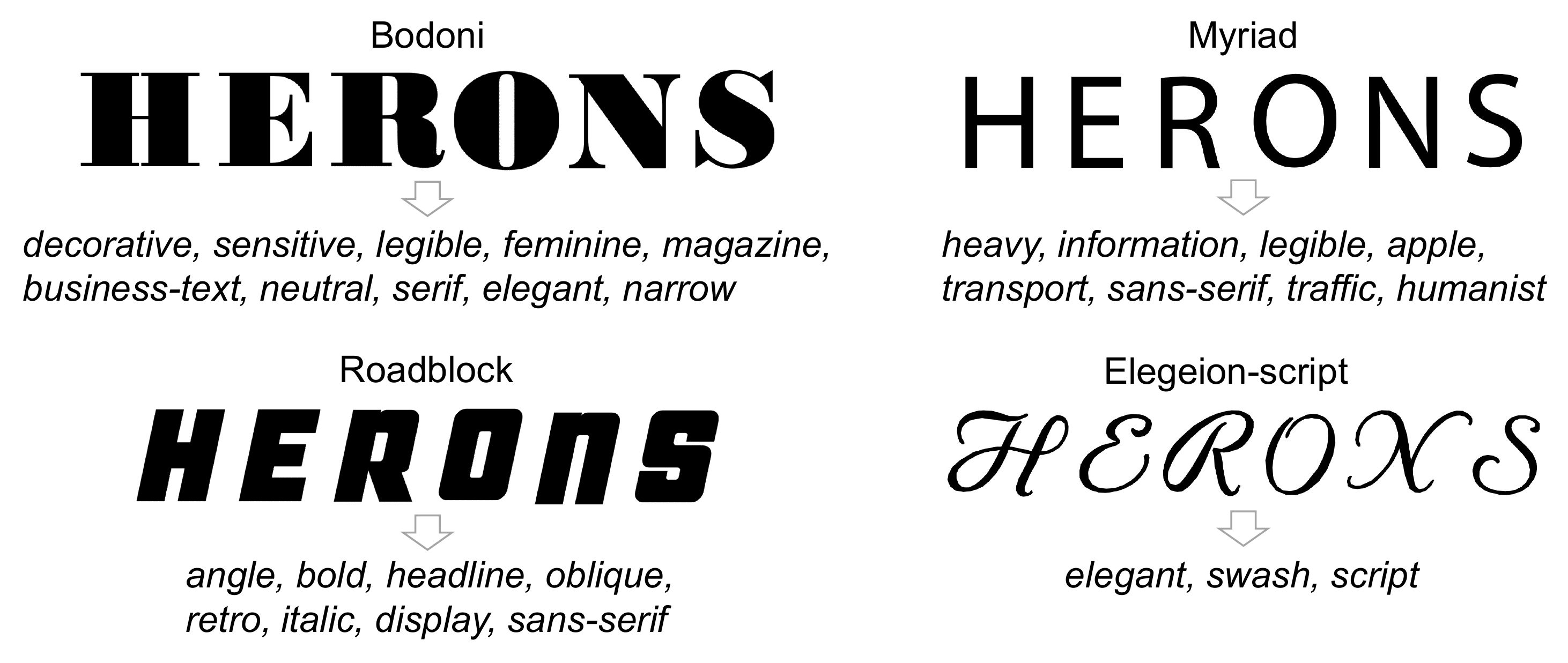}\\[-3mm]
    \caption{Fonts and their impressions (from MyFonts dataset~\cite{Chen2019large}). {\tt Bodoni}, {\tt Myriad}, etc. are font names.}
%    \vspace{-5mm}
    \label{fig:shape-and-impression}
\end{figure}
%---------------------------------
%---------------------------------
\begin{figure}[t]
    \centering
%    \vskip 5cm
    \includegraphics[width=1.0\linewidth]{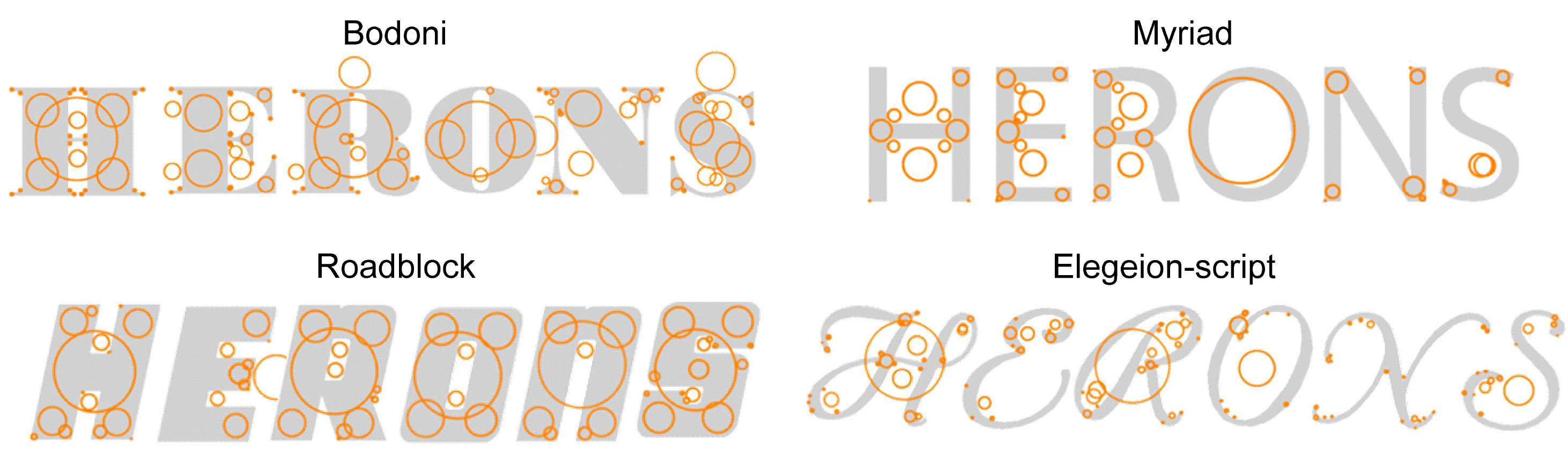}\\[-3mm]
    \caption{SIFT keypoints. The center of each circle represents the location of a keypoint and the radius represents the scale at the keypoint. Roughly speaking, the SIFT descriptor at each keypoint represents the local shape around the circle.}
%    \vspace{-5mm}
    \label{fig:SIFT}
\end{figure}
%---------------------------------
Different fonts, or typefaces, have different impressions. Each font just has a specific shape style and this style is converted to some special impression in our mind.  At {\tt MyFonts.com}, each font is tagged with several impression words. We can see
those words as a result of a shape-to-impression translation by humans.\par
It is still not clear why and how different font shape styles give different impressions. As will be reviewed in Section~\ref{sec:review}, many subjective experiments have proved that font shapes surely affect their impression, readability, and legibility. However, most of those experiments just confirm that a specific font (such as Helvetica) has a specific impression (such as reliable). In other words, they do not reveal more detailed relationships or general trends between font shapes and impressions.\par
To understand the shape-impression relationship, we need more objective experiments using a large-scale dataset. Fortunately, Chen et al.~\cite{Chen2019large} have published a shape-impression dataset by using the content of {\tt MyFonts.com}. This is a very large dataset containing 
18,815 fonts and 1,824 vocabularies of {\em impression tags} attached to each font. 
Fig.~\ref{fig:shape-and-impression} shows several examples of fonts and their impression tags. Some tags directly express 
font style types such as {\it Sans-Serif} and {\it Script}, more shape-related properties such as {\it Bold} and {\it Oblique} and more abstract impressions such as {\it Elegant} and {\it Scary}. 
\par
If we can reveal the relationships between font shapes and those impression tags, it will be very meaningful for not only several practical applications but also more fundamental research. Example practical applications include (1) font selection or recommendation systems that provide a suitable font according to typographer's ambiguous requests by impression. (e.g.,  \cite{Chen2019large,choi2018fontmatcher}) and (2) font generation systems that can accept the impression words as the constraints on the generation (e.g.,  \cite{matsuda-ICDAR2021}). From the viewpoint of more fundamental research, the shape-impression relationship is still unrevealed and will give important evidence to understand the human perception mechanism. \par
To understand those relationships between font shapes and impressions, one of the promising approaches is {\em part-based approach}, where a font image is decomposed into a set of local parts and then the individual parts and their combinations are correlated with impressions. This is because the shape of `A' is comprised of two shape factors --- the letter shape (so-called `A'-ness~\cite{hofstadter}) and the font style. The part-based approach can discard the letter shape, while retaining various impression clues from local shapes, such as serif, curvature, corner shape, stroke width, etc.\par
\begin{figure}[t]
  \hspace{-1cm}
  \begin{minipage}[b]{0.3\textheight}
    \centering
    \includegraphics[height=0.25\textheight]{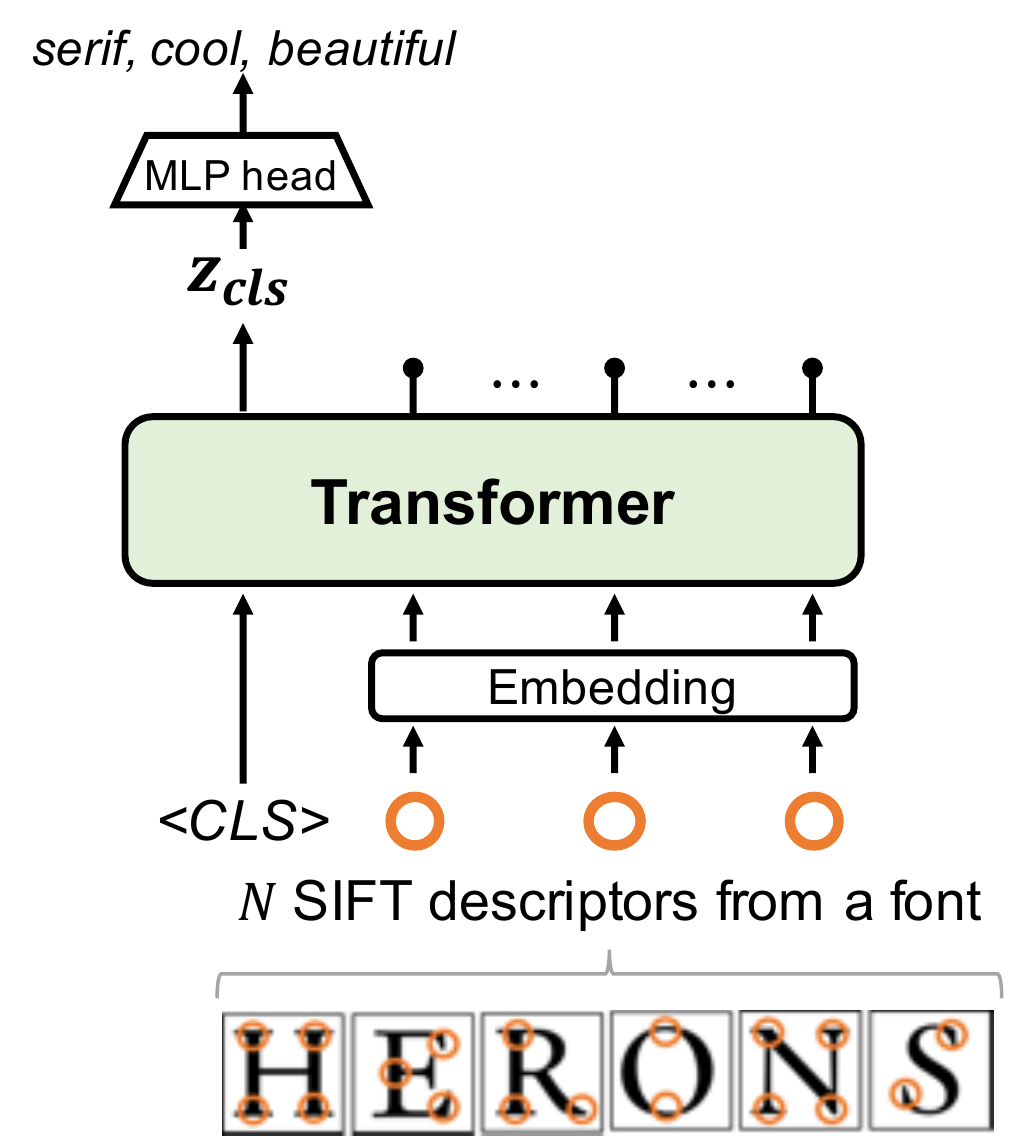}\\
    (a)~Multi-label classification.
  \end{minipage}
  \hspace{-5mm}
  \begin{minipage}[b]{0.3\textheight}
    \centering
    \includegraphics[height=0.252\textheight]{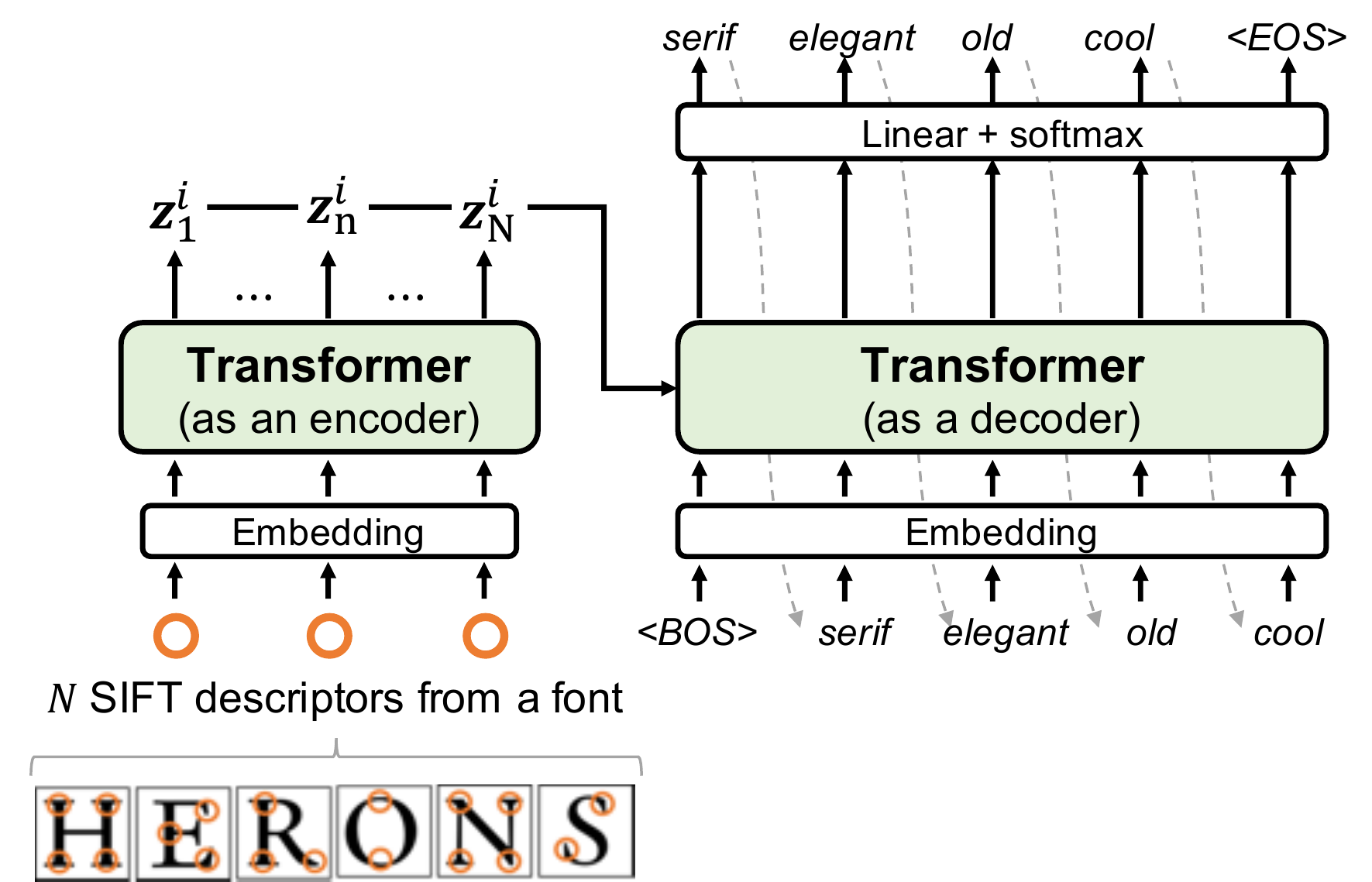}\\
     \hspace{2cm}(b)~Translation.
  \end{minipage}
  \caption{Two approaches of part-based shape-impression relationship analysis with Transformers.\label{fig:overall}}
\end{figure}

This paper proposes a novel method for part-based shape-impression analysis that fully utilizes Transformer~\cite{vaswani2017attention}, which is a recent but already well-known deep neural network architecture. We first train Transformer to output impressions for a given set of local shapes. Then, we analyze the trained Transformer in various ways to understand the important local shapes for a specific impression. The advantages of Transformer for our analysis are threefold. 
\begin{enumerate}
    \item Transformer is a versatile model and offers us two different approaches. As shown in Fig.~\ref{fig:overall}, the classification approach (a) accepts $N$ local descriptors by SIFT~\cite{SIFT} as its input elements and outputs the probability of each of $K$ impression classes. In the translation approach (b), one Transformer as an encoder accepts $N$ SIFT descriptors as its input and then encodes them into latent vectors (called ``key'' and ``value''), and then the latent vectors are fed to another Transformer as a decoder that outputs a set of impression words like a translation result.
    \item Transformer can accept a variable number of input elements. Since the number of local shapes from a single font image is not constant, this property is suitable for our task. 
    \item The most important advantage is its {\em self-attention} mechanism.  Self-attention determines a weight for every input element by considering other input elements. Therefore, if we input local shapes to Transformer as multiple input elements, their correlation is internally calculated and used for the task. For example in the classification approach, the correlation among the local shapes that are important for a correct impression class will be boosted through the self-attention mechanism.
\end{enumerate}
\par

%---------------------------------

We also introduce explainable-AI (XAI) techniques \cite{occlusion,sundararajan2017axiomatic} for a deep understanding of the importance of local shapes on a specific impression. In the experiment, we reveal the important parts for a specific impression by several different techniques.\par
Our contributions are summarized as follows:
\begin{itemize}
\item We propose two methods for part-based font-impression analysis using Transformer. They are a multi-label impression classifier and a shape-to-impression translator. To the authors' knowledge, this is the first attempt that utilizes Transformer for impression analysis. 
\item Using a large font-impression dataset, we experimentally prove that our Transformer-based methods enable us to realize more reliable and flexible analysis than the previous approaches.
\item Our proposed method reveals local parts that well describe a specific impression with the help of XAI. 
\end{itemize}

% =======================================================================
\section{Related Work\label{sec:review}}
\subsection{Subjective impression analysis of fonts}
The analysis of the relationships between font shapes and impressions has a long history from the 1920's~\cite{davis1933determinants,poffenberger1923study}. Shaikh and Chaparro~\cite{shaikl} defined the impression of 40 fonts by collecting answers from 379 people. In 2014, O'Donovan et al.~\cite{o2014exploratory} have published their dataset with 200 fonts with impression tags. In this pioneering work on data-driven impression analysis, impression tags were only 37 vocabularies. More recently,  Chen et al.~\cite{Chen2019large} published a far larger font-impression dataset by using {\tt MyFonts.com} and then proposed a font retrieval system using impressions as a query. As will be detailed in Section~\ref{sec:dataset}, about 20,000 fonts are annotated with about 2,000 impression tags in their dataset. Although it contains noisy annotations, its rich variation is useful for analyzing the relationship between font shape and impression. For example, Matsuda et al.~\cite{matsuda-ICDAR2021} utilize it for a font generation with specific impressions.

\subsection{Objective impression analysis of fonts\label{sec:old-ueda}}
To the best of our knowledge, only a single attempt~\cite{ueda-ICDAR2021} has been made so far for shape-impression analysis by using a large dataset and an objective methodology. Although it also takes a part-based approach like ours, it has still a large room for improvement. One of the most significant issues is that all local shape descriptors extracted from a letter image are treated {\em totally independently} in  the framework of DeepSets~\cite{deepsets}. More specifically, DeepSets converts individual local shape descriptors to discriminative feature vectors independently, and then just adds those vectors into a single vector. Compared to ~\cite{ueda-ICDAR2021}, ours utilizes Transformer which can deal with the combinatorial relationships among the local descriptors by its self-attention mechanism, and therefore, as shown by the later experimental results, realizes a more accurate analysis.
\par
% Completed by Uchida 22:40 Jan 18

\subsection{Transformer}
Transformer~\cite{vaswani2017attention} is a multi-input and multi-output network and was originally developed for various tasks in natural language processing (NLP). It can be used as an encoder and also a decoder in some NLP tasks~\cite{otter2021:survey}, such as language translation. It is also used as a classifier in some NLP tasks~\cite{GPT3,devlin2018bert}, such as sentence sentiment classification. Transformer has also been applied to image classification tasks~\cite{khan2021survey}. Vision Transformer (ViT)~\cite{dosovitskiy2020image} is the most well-known application of Transformer to image classification. In ViT, an input image is first divided into small patches and then fed to Transformer. In this research, we use ViT as one of the comparative methods. Transformer is also applied to image captioning~\cite{cornia2020m2}. Although Transformer has been employed in a vast number of applications from its development, its usefulness in impression analysis tasks has not been proved so far.\par
% Completed by Uchida  Jan 20

% =======================================================================
\section{Dataset\label{sec:dataset} and Local Descriptor}
%データセットの詳細
As the font-impression dataset, we used the MyFonts dataset provided by Chen et al.~\cite{Chen2019large}, 
which contains 18,815 fonts. As shown in Fig.~\ref{fig:shape-and-impression}, multiple impression tags are attached to each font by font experts and non-experts (including the customers of {\tt MyFonts.com}). The vocabulary of tags is 1,824. By following \cite{ueda-ICDAR2021}, we remove minor tags with less than 100 occurrences. Consequently, we use $M=18,579$ fonts and $K=483$ impression tags. In the following experiments, we randomly split the dataset into three font-disjoint subsets for train, validation, and test, with the ratio of 0.8, 0.1, and 0.1.\par
As noted in Section~\ref{sec:intro} and shown in Fig.~\ref{fig:shape-and-impression}, some impression tags express a typical font style type, such as  {\it Sans-Serif} and {\it Script}, or a more shape-related property, such as {\it Bold} and {\it Oblique}, or a more abstract impression, such as {\it Elegant} and {\it Scary}. There is no clear taxonomy among them, and we will simply refer to them as impression tags in this paper. \par
In this paper, we apply SIFT~\cite{SIFT} for extracting local shape descriptors from six letter images of `H,'`E,'`R,'`O,'`N,' and `S,'  as shown in Fig.~\ref{fig:SIFT}.
These letters are often employed in typographic work or font design because they contain almost all elements of English alphabets, such as horizontal strokes, vertical strokes, diagonal lines (`N' and `R'), intersections (`H' and `R'), curves (`R' and `S'), corners (`E' and `R'), and a circle (`O'). Using those six letters (instead of all the alphabets) can limit the number of SIFT vectors and improve the efficiency of training and testing Transformers.
\par
As local shape descriptors, several successors of SIFT have been proposed so far. For example, SURF was proposed as an efficient approximation of SIFT. BRIEF and ORB are also faster versions of SIFT and they provide binary features (for Hamming distance-based fast matching); namely, they do not represent the local shapes in a direct way. We, therefore, decided to use SIFT mainly, following \cite{ueda-ICDAR2021}.

% Completed by Uchida 20th

% =======================================================================
\section{Shape-Impression Relationship Analysis by Multi-label Classification Approach}
\label{sec:method1}
% =======================================================================

%-----------------------------------------------------------------------
\subsection{Transformer as a multi-label classifier}
For our purpose of the part-based shape-impression relation analysis, we use Transformer as a multi-label impression classifier. If Transformer is trained to estimate the impression tags for a set of local shape descriptors extracted from a font image, the trained Transformer should know some shape-impression relationships internally. We, therefore, can understand the local shapes that are important for a specific impression by visualizing and analyzing the trained Transformer. Moreover, this analysis result will reflect the combinatorial relationship among the local shapes because of the self-attention mechanism in Transformer. \par
Fig.~\ref{fig:overall}~(a) shows the multi-label impression classification by Transformer. Each of $N$ SIFT descriptors extracted from six letter images (as noted in Section~\ref{sec:dataset}) is first embedded into another feature space by a single fully-connected (FC) layer and then the resulting $N$ feature vectors and a dummy vector, called class-token \CLS, are fed into Transformer as $N+1$ inputs. Note that the number $N$ is variable with font types. 
After going through a transformer layer (which is comprised of self-attention and FC) $L$ times, we have the Transformer output corresponding to \CLS. This output is finally fed to a multi-layer perceptron (MLP)-head with a sigmoid function to have the probability of each of $K$ impressions.\par

The above model is similar to ViT~\cite{dosovitskiy2020image} but very different at three points. 
First, ours input local descriptors (which are extracted irregularly like Fig.~\ref{fig:SIFT}), instead of regular square patches. Second, ours does not employ the position encoding of the input elements. 
This is because the locations of the SIFT keypoints where local descriptors are extracted will incur the letter shape (i.e., `A'-ness~\cite{hofstadter}) and therefore disturb our part-based impression analysis. Without the position encoding, $N$ SIFT descriptors are treated as a set with $N$ elements; consequently, we do not need to pay attention to the order of the $N$ features when we input them to Transformer. The third and a rather minor difference is that ours have a variable number $N$ of inputs, whereas ViT always accepts the same number (i.e., the number of patches).
% completed by Uchida 20th
%-----------------------------------------------------------------------
\subsection{Implementation details}
The multi-label classification model in Fig.~\ref{fig:overall}~(a) is comprised of five transformer layers internally (i.e., $L=5$) . The self-attention module is organized as so-called multi-head attention with five heads. Each 128-dimensional descriptor input is embedded into 128-dimensional space by an FC layer. The maximum number of the input descriptors is 300. If more than 300 keypoints are detected from the six letters (``HERONS''), 300 descriptors are randomly selected. Otherwise, the zero-padding is applied to have 300 inputs in total. MLP head accepts the output of the Transformer and outputs a $K$-dimensional class likelihood vector via an FC layer. Binary Cross Entropy (BCE) is used as the multi-label classification loss. The Adam optimizer~\cite{kingma2014adam} is used with a learning rate of 0.001.  \par
% Checked by Uchida 20th

%---------------------------------
\begin{figure}[t]
    \centering
%    \vskip 5cm
    \includegraphics[width=1.0\linewidth]{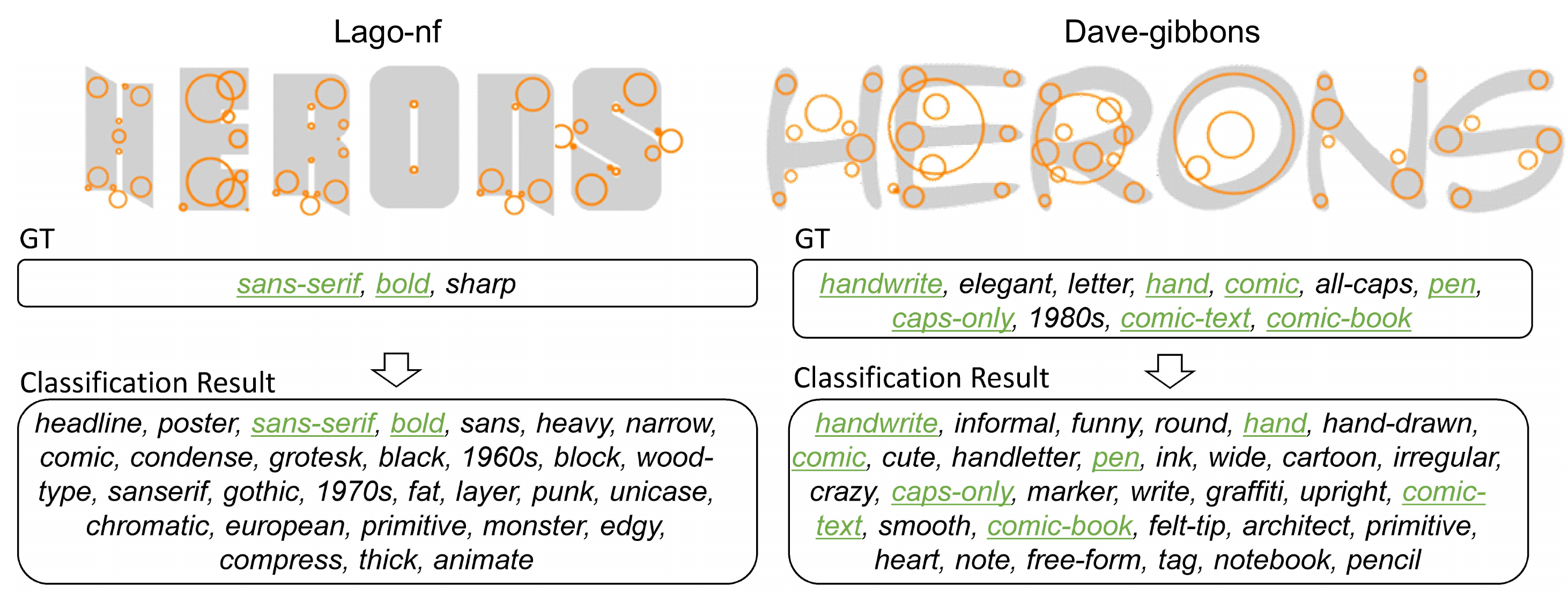}\\[-2mm]
    \caption{Example results of multi-label classification. Two fonts are from the test set.}
%    \vspace{-5mm}
    \label{fig:multi-label-exmaples}
\end{figure}
%---------------------------------
%-----------------------------------------------------------------------
\subsection{Classification examples\label{sec:classificaiton-examples}}
Fig.~\ref{fig:multi-label-exmaples} shows two test examples of the multi-label classification. Many impressions in the ground-truth are correctly found in the result. Both classification results include significantly more impressions than the ground-truth. A closer look at those excessively detected impressions will find that they are often very similar to one of the labeled impressions. For example, {\it cartoon} and {\it pencil} for {\tt Dave-gibbons} are similar to the labeled impressions {\it comic} and {\it pen}, respectively. The impression labels of the MyFonts dataset are attached by unspecified people and there is no guarantee that all possible impressions among $K$ vocabularies are fully attached and organized. This situation is the so-called {\it missing labels} condition. 
The impression {\it cartoon} and {\it pencil} might be missing labels; they could be included in the ground-truth, but none attached them. 
% Checked by Uchida Jan 19 0:15

%-----------------------------------------------------------------------
\subsection{Shape-Impression relation analysis by group-based occlusion sensitivity\label{sec:occlusion-sensitivity}}
To understand the importance of local shapes on a specific impression, we analyze the trained Transformer by using a classical but reliable explainable-AI (XAI) technique, called {\em occlusion sensitivity}. Occlusion sensitivity was first introduced to an image recognition task in~\cite{occlusion}. Its idea is to measure the change of the class likelihood (BCE for our case) when an input part is removed, i.e., occluded. If the removal of an input part drastically decreases the likelihood of a certain class, we consider that the part is very important for the class. In~\cite{occlusion}, a gray image patch is superimposed on the input image to remove  a part.
\par
In our case, the naive application of occlusion sensitivity is to remove a local descriptor from the $N$ inputs and observe the change of the class likelihood. However, this naive application is not meaningful; this is because removing a descriptor from the $N$ inputs often does not affect the likelihood, because very similar descriptors exist in the remaining $N-1$. In other words, if there are similar descriptors, effective occlusion is no longer possible by this one-by-one removal.
\par
We, therefore, group the $N$ descriptors (i.e., local shapes) into $Q$ clusters and remove all the descriptors belonging to a certain cluster $q\in [1, Q]$, from the $N$ descriptors. By this {\em group-based occlusion sensitivity}, we can understand the importance of a particular type of local shape for a specific impression. For each of $K$ impressions and each of $Q$ local shape types, we measure this sensitivity and then have a $Q$-dimensional sensitivity vector for each impression. For clustering, $k$-means is performed on all SIFT descriptors extracted from all the training font images. The $Q$ centroids are treated like ``visual words'' and SIFT descriptors extracted from a certain font image are quantized into $Q$ types.
% Checked by Uchida Jan 19  AM0300
% =======================================================================
\section{Shape-Impression Relationship Analysis by Translation Approach}
\label{sec:method2}
% =======================================================================
\subsection{Transformer as a shape-to-impression translator}
Inspired by the fact that cascading two Transformers realizes a language translator~\cite{vaswani2017attention}, we now consider another shape-impression relation analysis approach with Transformers. Fig.~\ref{fig:overall}~(b) shows the overall structure of the shape-to-impression translator, which roughly follows the structure of \cite{vaswani2017attention}. Two Transformers work as an encoder and a decoder, respectively. In our task, we expect that the decoder outputs the correct impressions by feeding a set of $N$ SIFT descriptors to the encoder.\par   
\par
The encoder is the same as the multi-label classifier of Fig.~\ref{fig:overall}~(a), except it has $N$ outputs and no class-token input. More precisely, after embedding $N$ SIFT descriptors into a certain feature space by an FC layer, the resulting $N$ vectors are fed to the encoder Transformer. Then, the encoder outputs $N$ corresponding latent vectors. Note that the $n$-th latent vector conveys the information from not only the $n$-th SIFT descriptor but also the remaining $N-1$ SIFT descriptors, because of the self-attention mechanism in the encoder.
\par
By utilizing the $N$ latent vectors, the decoder Transformer outputs impression words one by one. More precisely, the special token \BOS (beginning-of-sentence) is first input to the decoder and then the decoder outputs the first $(K+3)$-dimensional impression likelihood vector. The extra three dimensions correspond to three special tokens, \BOS, \EOS (end-of-sentence), and \PAD\footnote{\PAD token is used when we train the decoder. {\PAD} tokens are added to the end of the ground-truth (i.e., the sequence of the labeled impressions) multiple times until the length of the ground-truth reaches the maximum output length.} (padding), respectively. The impression with the maximum likelihood in the $K$ vector elements is determined as the first impression. Second, the first impression word is fed to the decoder as the second input, and then the second impression is output. By repeating this process until \EOS  is output, we have a sequence of impressions.\par 
% Checked by Uchida Jan 20
%-----------------------------------------------------------------------
\subsection{Implementation details}
For training the decoder of the shape-to-impression translator, we need to specify the unique order of impressions as its ideal output, although they have no pre-defined order. We, therefore, rank the $K$ impressions according to their frequency; the most popular impression is the top and the least popular impression is the bottom. Consequently, the decoder tends to outputs impressions in the descending order of their popularity. \par
% Checked by Uchida Jan 20
%
Both Transformers in Fig.~\ref{fig:overall}~(b) are organized by five transformation layers ($L=5$). The self-attention module is organized by single-head attention. The embedding process for the $N$ input descriptors, the optimizer, and the learning rate are the same as for the multi-label classifier in the previous section. Only for the input of the decoder, the sinusoidal position encoding~\cite{vaswani2017attention} is used. We employ the beam search during translation. Cross-Entropy is used as the loss function.
% Checked by Uchida Jan 20
%-----------------------------------------------------------------------
\subsection{Translation examples}
\begin{figure}[t]
    \centering
%    \vskip 5cm
    \includegraphics[width=1.0\linewidth]{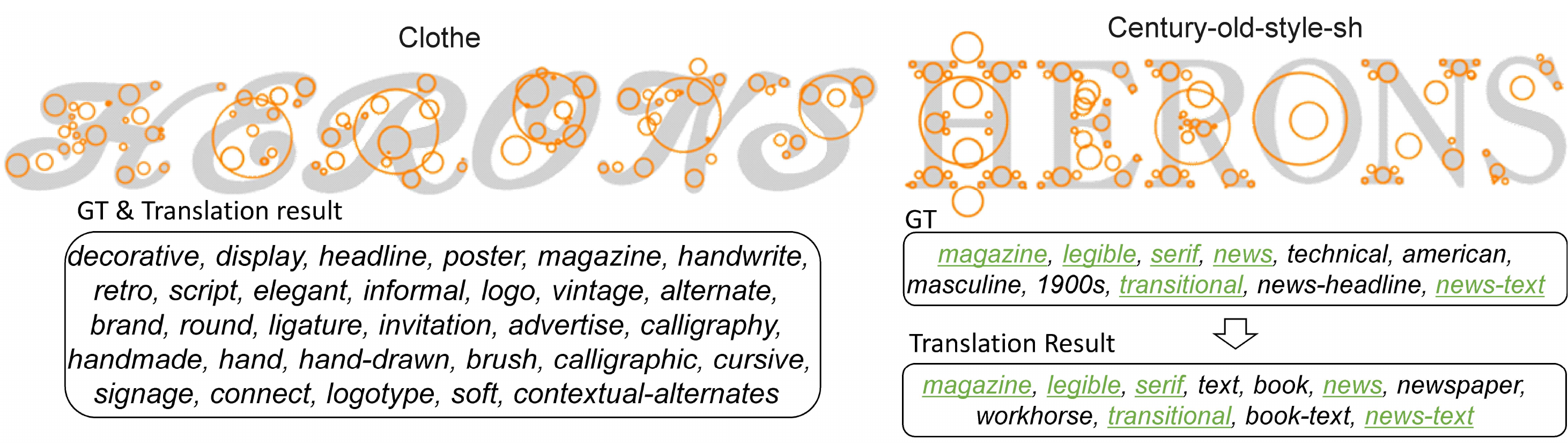}\\[-3mm]
    \caption{Example results of shape-to-impression translation. Two fonts are from the test set.}
%    \vspace{-5mm}
    \label{fig:translation_example}
\end{figure}

Fig.~\ref{fig:translation_example} shows two test examples of the shape-to-impression translator. Surprisingly, for the test font {\tt Clothe}, the translator could output all 30 labeled impressions perfectly. For {\tt Century-old-sty}, the translator cannot give the perfect result. As noted in Section~\ref{sec:classificaiton-examples}, the ground-truth of the MyFonts dataset often has missing labels and therefore some impressions similar to a labeled impression are often detected excessively. 
% Checked by Uchida 20

%---------------------------------
%---------------------------------

%-----------------------------------------------------------------------
\subsection{Shape-impression relation analysis using integrated gradients\label{sec:ig-analysis}}
To evaluate the shape-impression relationship with the translation approach, we use a gradient-based XAI method, called {\em Integrated Gradients} (IG)~\cite{sundararajan2017axiomatic}, which has a high versatility for explaining arbitrary networks and has thus been recently utilized in various applications\cite{cornia2020m2,he2019towards}. Although our translation approach employs a complex encoder-decoder framework, IG is still applicable to it for understanding its input (local shape) and output (impression) relationship.\par
More precisely, in our case, IG is used to evaluate the decoder output gradients against one of $N$ local shapes. The key idea of IG is to evaluate multiple gradients that are measured by gradually changing the input vectors from a zero vector (called ``baseline'' in \cite{sundararajan2017axiomatic}) to the original SIFT descriptors. The multiple gradients are then aggregated (i.e., integrated) into the integrated gradient, i.e., IG. If an input local shape has a large IG to a specific impression, the input is important for the impression. 
% Checked by Uchida 20
% =======================================================================
\section{Experimental Results}
% =======================================================================
\subsection{Quantitative evaluation of the trained Transformer}
Although our main purpose is to analyze the relationship between local shapes and impressions through the trained Transformers, we first need to confirm that those Transformers show reasonable performance in the multi-label classification task and the translation task. In other words, if the trained Transformers' performance is poor on these tasks, the relationship learned in the Transformers is not reliable. We, therefore, first conducted a quantitative evaluation of the trained Transformers.\par
% Checked by Uchida Jan 20

\subsubsection{Evaluation metrics} 
\begin{description}
\item{\bf F1@100, F1@200, and F1@all} are evaluation metrics for multi-label classification tasks. They are simply the average of the F1 scores of the most frequent 100, 200, and all $K(=483)$ impressions, respectively. For the classification approach of Section~\ref{sec:method1}, the multi-label classification is first made by applying the threshold $0.5$ to  the class likelihood values of $K$ impressions. If the likelihood of the $k$-th impression class is larger than $0.5$,
we determine that the set of $N$ local shapes show the impression $k$. Then, the F1 score is calculated for $K$ individual impressions and then finally averaged. For the translation approach of Section~\ref{sec:method2}, we calculate the $K$ F1 scores by comparing the output impression sequence (i.e., the translation result) to the correct set of impressions for each font.
\item{\bf mean average precision (mAP)} is also an evaluation metric for multi-label classification tasks. It is the average of the average precisions of all $K$ impressions. The average precision for the $k$-th impression is calculated by using the list of all $M$ fonts ranked in the descending order of the likelihood of the $k$-th impression. More precisely, it is calculated as $(\sum_{m=1}^{M_k}m/r_m)/M_k$, where $M_k$ is the number of fonts labeled with $k$, $r_m\in [1, M]$ is the rank of the font with the $m$th largest likelihood among $M_k$. Since the translator approach does not provide the impression likelihood, we cannot calculate its mAP.
\end{description}
% Checked by Uchida 20

%---------------------------------
%\begin{figure}[t]
%    \centering
%    \vskip 5cm
%   \includegraphics[width=1.0\linewidth]{figs/metrics.pdf}\\[-3mm]
%    \caption{Each metrics of methods.}
%    \vspace{-5mm}
%    \label{fig:metrics}
%\end{figure}
%---------------------------------
%---------------------------------
\begin{table}[t]
    \centering
%    \vskip 5cm
    \caption{Quantitative evaluation result.}
    \label{tab:comparison}
    \begin{tabular}{|c|c|c|c|c|}
    \hline
         & Multi-label classification & Translator & ViT~\cite{dosovitskiy2020image} & DeepSets~\cite{ueda-ICDAR2021} \\ \hline
    Inputs & SIFT & SIFT & patch & SIFT \\
    \hline\hline
    F1@100$\uparrow$ & \textbf{0.301} & 0.264 & 0.264 & 0.279 \\
    \hline
    F1@200$\uparrow$ & \textbf{0.221} & 0.186 & 0.185 & 0.194 \\
    \hline
    F1@all$\uparrow$ & \textbf{0.145} & 0.117 & 0.109 & 0.110 \\
    \hline
    mAP $\uparrow$& \textbf{0.135} & N/A & 0.115 & 0.115 \\
    \hline
    \end{tabular}
\end{table}
%---------------------------------
\subsubsection{Results and comparisons}
In Table~\ref{tab:comparison}, the impression estimation accuracies of the proposed two approaches are compared with the performance by two existing methods, ViT\cite{dosovitskiy2020image} and DeepSets\cite{ueda-ICDAR2021}. This table clearly shows that our classification approaches with local shape inputs outperform the existing methods. Comparison between our classification approach and ViT indicates the SIFT descriptors\footnote{We have tried the SURF descriptors instead of the SIFT descriptors to show the justification to select SIFT as local shape descriptors. We found no significant differences between them. More precisely, the multi-label classifier using SURF achieved about 0.16-point higher mAP and 0.05-point lower F1@all than SIFT.} are more suitable than the regular patches, for capturing the important local shapes. More importantly, the fact that our classification approach outperforms DeepSets (with the SIFT descriptors) suggests that the correlation among local shapes by the self-attention in Transformer is important to estimate the font impressions. (As noted in Section~\ref{sec:old-ueda}, DeepSets treats local shapes totally independently.) The translator shows a similar performance with the existing methods. 

%---------------------------------
\begin{figure}[t]
    \centering
%    \vskip 5cm
    \includegraphics[width=1.0\linewidth]{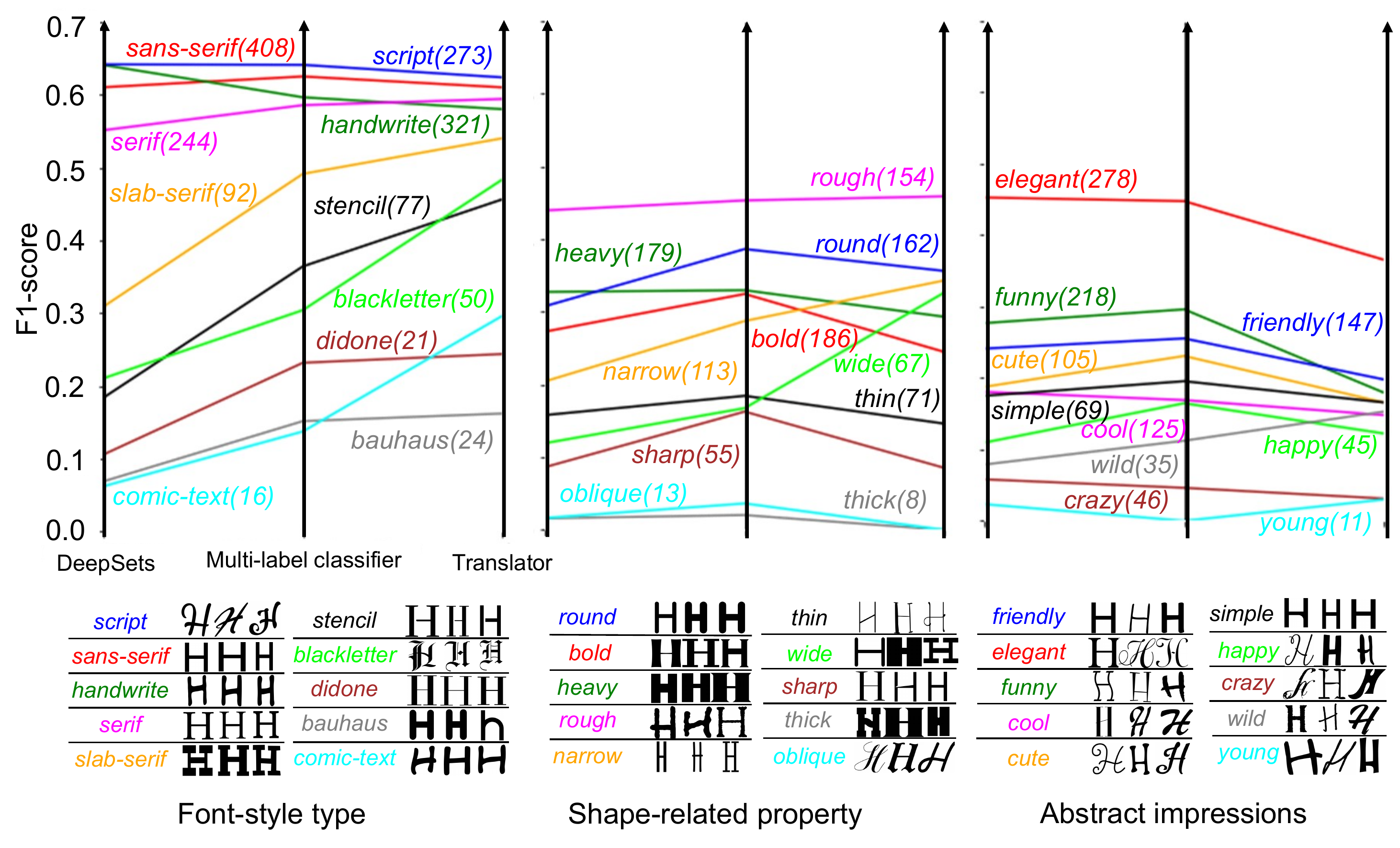}\\[-3mm]
    \caption{F1 scores of three methods for 30 impressions are represented by parallel coordinate plots. The number of the test fonts with the impression is parenthesized. Three font image examples for each impression are also shown.}
%    \vspace{-5mm}
    \label{fig:parallelcoordinate}
% \medskip
%     \centering
% %    \vskip 5cm
%     \includegraphics[width=0.9\linewidth]{figs/impression_example.pdf}\\[-3mm]
%     \caption{Font image examples for each impression.}
%     \vspace{-5mm}
%     \label{fig:impression_example}
\end{figure}
%---------------------------------
Fig.~\ref{fig:parallelcoordinate} shows the F1 scores of three methods (DeepSets-based method~\cite{ueda-ICDAR2021}, the multi-label classifier, and the shape-to-impression translator) for 30 impressions by parallel coordinate plots. The 30 impressions are selected from font-style type impressions (such as {\it sans-serif}), shape-related property impressions (such as {\it round}), and abstract impressions (such as {\it elegant}). Fig.~\ref{fig:parallelcoordinate} also shows examples of font images.\par
Although it is difficult to find strong trends from Fig.~\ref{fig:parallelcoordinate}, several weak trends can be found as follows. (1)~The multi-label classifier shows higher F1 than (or equal F1 to) the DeepSets-based method for most impressions.
(2)~The multi-label classifier shows significantly higher F1 for the impression whose discriminative shape needs to be described by a combination of local parts. Multiple narrow gaps (with abrupt stoke ends) of \textit{stencil}, square-shaped serifs of \textit{slab-serif}, and densely-distributed (sparsely-distributed) local parts of \textit{narrow} (\textit{wide}) are examples. This might be because of the positive effect of self-attention in Transformer. (3)~Frequent impressions tend to get higher F1 values and those values are rather stable among the three methods.\par
Although it is also difficult to find general trends in the relationship between the multi-label classifier and the translator, the latter often outperforms the former for minor but strong impressions, such as \textit{blackletter} and \textit{comic-text}. This will be because the fonts with such strong impressions tend to have a stable impression set. In fact, the stability is beneficial for the translator. Since the translator recursively outputs impressions in order of popularity, major but unexpected impressions will interfere with the output of subsequent minor impressions. Therefore, if a font has a stable impression set, the translator can output minor but specific impressions, such as \textit{blackletter}. \par

\subsection{Analysis results of the shape-impression relationship}
\subsubsection{Analysis with the multi-label classification approach} 
%---------------------------------例
\begin{figure}[t]
    \centering
%    \vskip 5cm
    \includegraphics[width=1.0\linewidth]{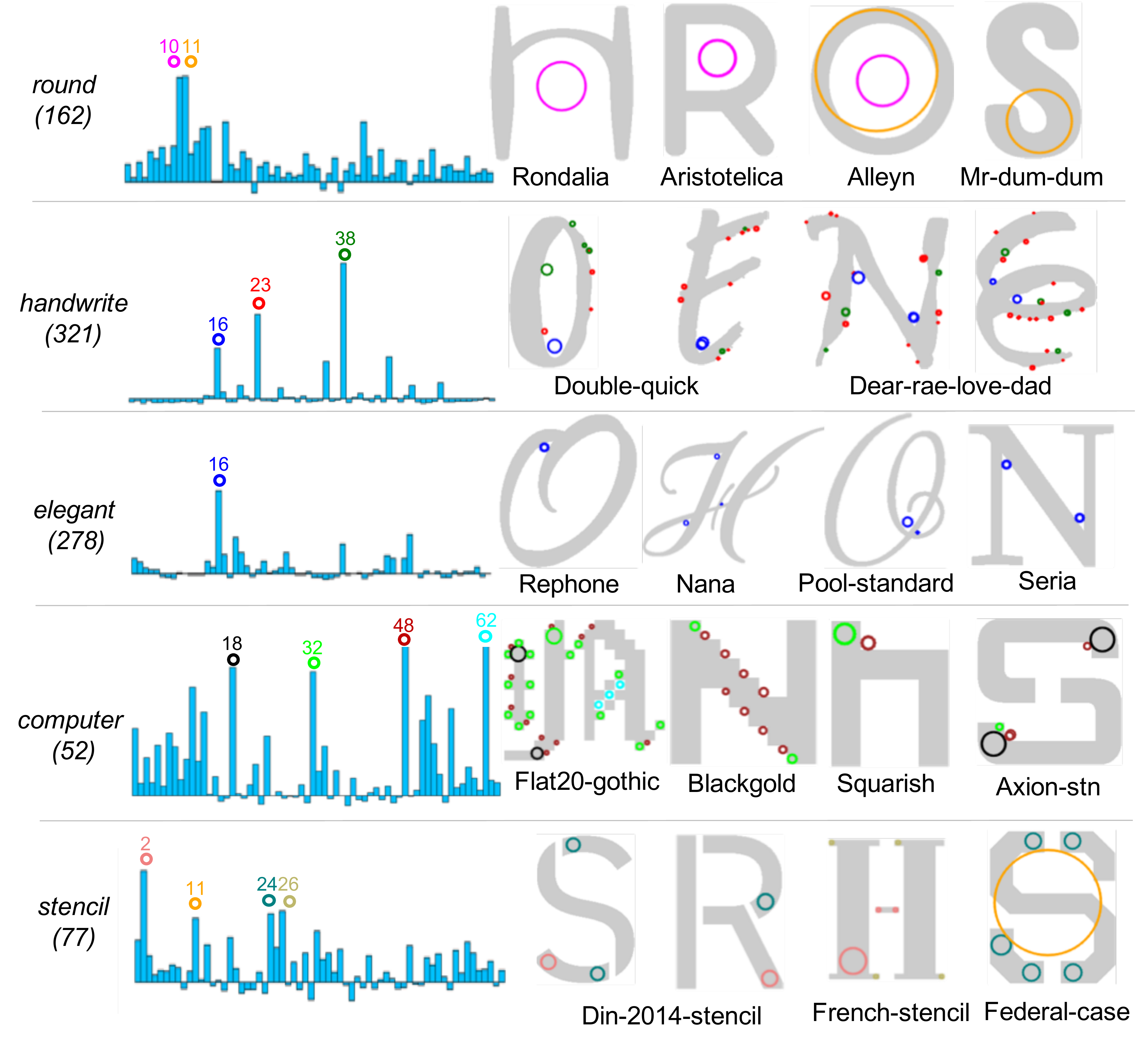}\\[-3mm]
    \caption{Important parts detected by the multi-label classification approach, for four impressions. The font names is shown below its image. The parenthesized number is the number of fonts with the impression.}
    \label{fig:important-parts-app1}
    \vspace{-3mm}   
\end{figure}
%---------------------------------
Fig.~\ref{fig:important-parts-app1} shows the local shapes that are important for specific impressions, such as {\it round} and {\it elegant}, by the multi-label classifier with the group-based occlusion sensitivity. The bar charts illustrate the sensitivity (i.e., the importance) of $Q=64$ local shape types for each of the four impressions. Since these bar charts show the {\em difference} from the median sensitivity, they have a minus element at the $q$th bin when the $q$th local shape type is less frequent than usual in the impression. Large positive peaks in the bar chart indicate the very important types. For example, for {\it round}, the 10th and 11th shape types are the most important. On the font images (from the test set), the local parts belonging to the important types are superimposed as circles, whose colors correspond to the types.\par
% Checked by Uchida 20
%
The following provides a brief interpretation of individual results. \textit{Round} has its peaks at $q=10,11$ that correspond to round corners or round spaces.
\textit{Handwrite} has a peak at $q=16$ that corresponds to sharp and asymmetric (i.e., organic) curves and two peaks at $q=23,38$ that correspond to rough shapes mimicking brush stroke.
\textit{Elegant} also has a peak at $q=16$. However, it does not have peaks at $q=23,38$; this indicates that \textit{Elegant} is organized by some organic and non-rough (i.e., smooth) curved strokes.
\textit{Computer} shows several peaks; among them the local shape types of $q=32, 48$ are often found together. This indicates that self-attention could successfully enhance the co-occurrence of those types for \textit{Computer}.
\textit{Stencil} has several peaks at $q=2,24,26$ that correspond to 
the abrupt end of a constant-width stroke. Since \textit{Stencil} often 
contains the abrupt-end shapes (for mimicking the actual stencil letters), 
excessive existence of those local shapes is important for its impression.

%hese shapes are considered to be very important for \textit{Stencil}. And \textit{Stencil} has also a peak at $q=11$ that corresponds to round space. This indicates that stencil-style has legibility.
%↑説明
%stencilはフォントスタイルタイプであり，線分が非連続であることが特徴的である．
%stencilでは2,24,26番のような平行線から成るストロークの端の形状が重要であることが分かった．
%stencilのフォントはストロークがつながっていない形状の為，2番のような平行線から成るストローク端が発生しやすい
%平行線の線分端はどんなフォントにも出やすいがself-attentionによって形状間の関係が考慮された結果，これらの形状がstencilにとって非常に重要な形状になったと考えられる

%
\subsubsection{Analysis with the translator} 
%---------------------------------
\begin{figure}[t]
    \centering
%    \vskip 5cm
    \includegraphics[width=1.0\linewidth]{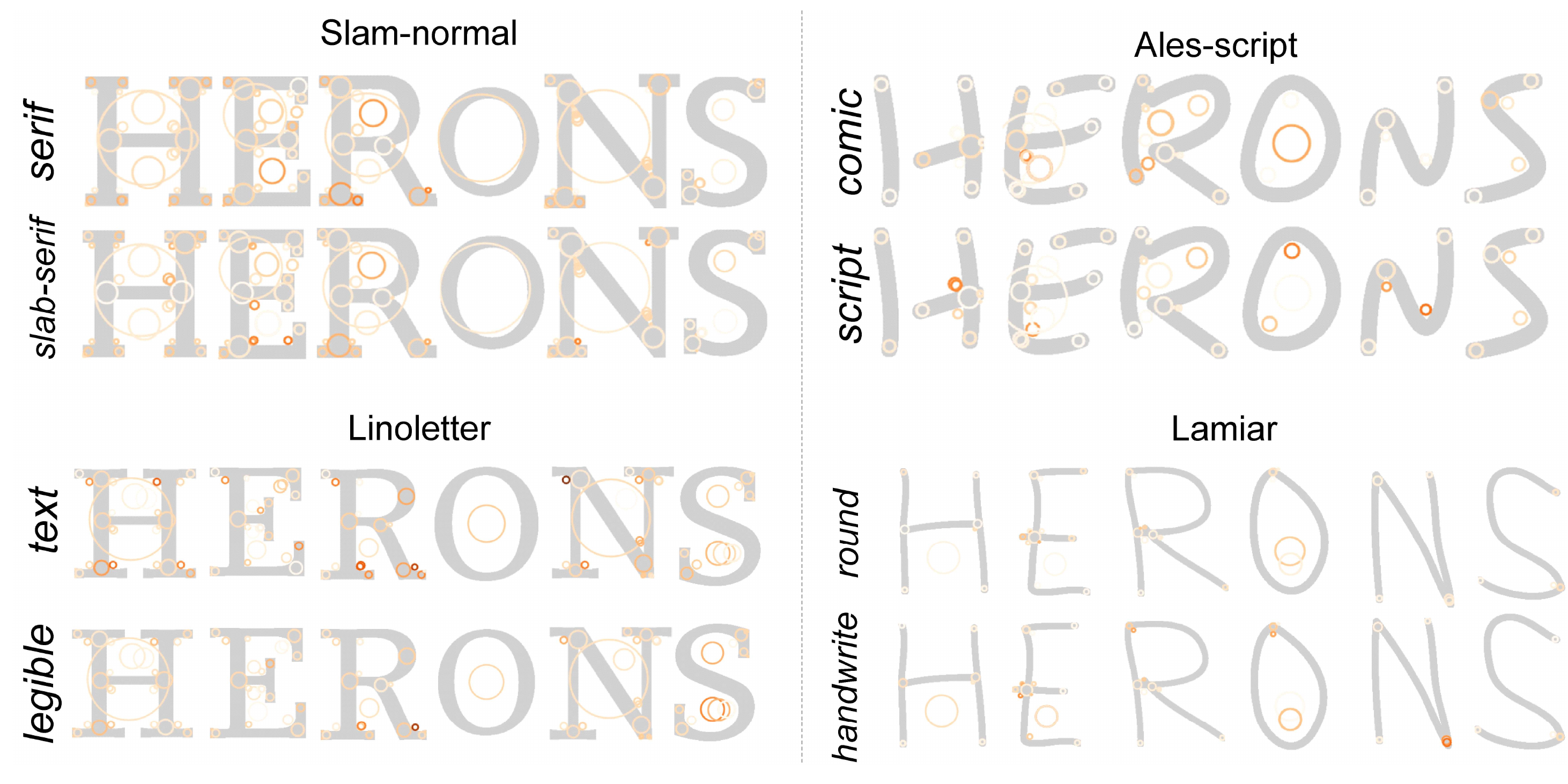}\\[-3mm]
    \caption{Important local parts by the translator with IG. More important parts are darker in color.}
    \vspace{-5mm}
    \label{fig:ig_result}
\end{figure}
%---------------------------------
Fig.~\ref{fig:ig_result} shows important local parts given by IG on two translation results. Compared to the multi-label classifier, the shape-to-impression translator can explain its important parts without any quantization and grouping by using IG. The darker circle has more IG, i.e., more important. In {\tt Slam-normal}, the serif parts of `R' contributes to {\it serif} and {\it slab-serif}. In {\tt Ales-script}, the important local parts are different according to  the impressions. For example, {\it comic-text} needs a large space of `O' and `R,' whereas {\it script} needs sharp curves of `N' and `O.' These results show that the shape which seems to be unique for a specific impression is important to translate into them.
 In {\tt Linoletter}, many of serif parts contribute to {\it text}, and the round space of `S' contributes to {\it legible}. In {\tt Lamiar}, the enclosed area of `O' is important for {\it round} and the sharp stroke ends of `E,' `R,' and `O' are important for {\it handwrite}. These results coincide with the results by the multi-label classifier shown in Fig.~\ref{fig:important-parts-app1}.
%lamiar->roundは'O'の中央から下の方にfig7で観測されたような空白が貢献していた．
%また，handwriteは'N','R','O'の鋭い曲線が貢献しており，fig7のq=16のような形状が貢献していた．
%linoletterではtext->serif部分，legible->'S'の曲線空白が貢献していた．またセリフ部分も赤く光っており，セリフ部分が読みやすさに貢献していることも考えられる．

%IG説明
%Fig.~\ref{fig:ig_result}に，section5.4で説明したように，translator approachでの翻訳結果に対する，各入力形状の寄与を可視化した結果を示す．
%図では，翻訳結果の印象にどの形状が寄与するかが，カラーリングによって図示されている（濃い方が貢献が大きい）．
%例えば{\tt Slam-normal}に着目すると，{\it serif}の翻訳に対しては，Rのセリフ部分の形状が翻訳に寄与している．
%また，{\it slab-serif}に対しては，Eに存在するslab-serifに特徴的な直角形状が寄与している．
%同様に，{\tt Ales-script}の例でも，印象特有であると考えられる形状（{\it comic-text}は丸みのある空白, {\it script}は急な曲線）の翻訳への寄与を確認した．
%この結果は，印象に重要であろう形状がそれらの翻訳出力のために重要であることを表す．

%また，translatorでは意味が類似する印象はともに出力される傾向にあるため，これらの寄与形状は，それらの印象全てにとっても同様に重要であると考えられる．
%例えば，{\it comic}は，{\it comic-text}や{\it handwrite}と出力される傾向があり，これら二つの印象にとっても丸みのある空白形状が重要であると考えられる．

% =======================================================================
\section{Conclusion and Future Work}
In this paper, part-based font-impression analysis is performed using Transformer. The versatility of Transformer offers us to realize two analysis methods: a multi-label impression classifier and a shape-to-impression translator. Using a large font-impression dataset, we experimentally prove that
the multi-label classifier could achieve better impression estimation performance; this means it can learn the trends between local shapes and impressions more accurately. We also revealed important local parts for specific impressions by using the trained Transformer with explainable-AI techniques.\par
Future work will focus on practical applications of the proposed methods to font selection or recommendation systems, font generation systems, and so on. Our analysis results on shape-impression relationships will be validated by collaborations with experts of cognitive psychology.  




% LNCS Homepage, \url{http://www.springer.com/lncs}. Last accessed 4
% \end{thebibliography}
\bibliography{das}
\bibliographystyle{splncs04}
\end{document}